\pdfoutput=1

\documentclass[11pt, dvipsnames]{article}

\usepackage[final]{acl}

\usepackage{times}
\usepackage{latexsym}

\usepackage[T1]{fontenc}

\usepackage[utf8]{inputenc}

\usepackage{microtype}

\usepackage{inconsolata}

\usepackage{graphicx}
\usepackage{multirow}    
\usepackage{array}       
\usepackage{caption}     
\usepackage{longtable}
\usepackage{adjustbox}
\usepackage{tabularx}
\usepackage{booktabs}   
\usepackage{makecell} 
\usepackage{forest}
\usepackage{listings}
\usepackage{xcolor}
\usepackage{fvextra}
\usepackage{todonotes}
\usepackage{subcaption}
\usepackage[breakable,skins]{tcolorbox}
\usepackage{amssymb}
\usepackage{pifont}

\newcommand{\DATASET}{\texttt{WikiMixQA}}

\newcommand{\ie}{\textit{i.e.}}
\newcommand{\eg}{\textit{e.g.}}

\newcommand{\xmark}{\textcolor{red}{\ding{55}}}   

\definecolor{darkgreen2}{HTML}{009B55}
\definecolor{myblue2}{HTML}{29abe2}
\definecolor{myorange2}{HTML}{f7931e}

\definecolor{mypurple2}{HTML}{9823FF}

\newif\ifcomments
\commentstrue

\definecolor{myred}{HTML}{E11D3F}
\definecolor{myorange}{HTML}{F04A00}
\definecolor{myblue}{HTML}{1974D2}
\definecolor{mypurple}{HTML}{6A0DAD}
\definecolor{NFpurple}{rgb}{0.8,0.0,0.8}
\ifcomments

\else
    \excludecomment{todoenv}
    \excludecomment{noteenv}
    \newcommand{\todo}[1]{}
    \newcommand{\db}[1]{}
    \newcommand{\dbm}[1]{}
\fi

\title{\DATASET: A Multimodal Benchmark \\for Question Answering over Tables and Charts}

\author{
  Negar Foroutan\textsuperscript{1}, Angelika Romanou\textsuperscript{1}, Matin Ansaripour\textsuperscript{1} \\ \textbf{Julian Martin Eisenschlos\textsuperscript{2}\textsuperscript{3}},
  \textbf{Karl Aberer\textsuperscript{1}}, \textbf{Rémi Lebret\textsuperscript{1}}  \\
     \textsuperscript{1}EPFL,
 \textsuperscript{2}Google DeepMind, \textsuperscript{3} Universidad Nacional de Córdoba
    \\
 \small{
   \textbf{Correspondence:} \href{mailto:negar.foroutan@epfl.ch}{\{negar.foroutan\}@epfl.ch}
 }}

\begin{document}
\maketitle
\begin{abstract}

Documents are fundamental to preserving and disseminating information, often incorporating complex layouts, tables, and charts that pose significant challenges for automatic document understanding (DU). While vision-language large models (VLLMs) have demonstrated improvements across various tasks, their effectiveness in processing long-context vision inputs remains unclear.
This paper introduces \DATASET, a benchmark comprising 1,000 multiple-choice questions (MCQs) designed to evaluate cross-modal reasoning over tables and charts extracted from 4,000 Wikipedia pages spanning seven distinct topics. Unlike existing benchmarks, \DATASET{} emphasizes complex reasoning by requiring models to synthesize information from multiple modalities.
We evaluate 12 state-of-the-art vision-language models, revealing that while proprietary models achieve $\sim$70\% accuracy when provided with direct context, their performance deteriorates significantly when retrieval from long documents is required. Among these, GPT-4-o is the only model exceeding 50\% accuracy in this setting, whereas open-source models perform considerably worse, with a maximum accuracy of 27\%. These findings underscore the challenges of long-context, multi-modal reasoning and establish \DATASET{} as a crucial benchmark for advancing document understanding research.\footnote{Code and dataset are released \href{https://github.com/negar-foroutan/WikiMixQA}{here}.}
\end{abstract}

\section{Introduction}
Documents serve as a fundamental medium for preserving and exchanging information, with millions being generated daily across various domains. Beyond plain text, they often include complex layouts, tables, and images, making automatic document understanding (DU) a crucial challenge. NLP-driven DU enables efficient extraction, organization, and interpretation of this information, supporting real-world information retrieval and decision-making.

\begin{figure}[ht]
    \centering
    \includegraphics[width=\linewidth]{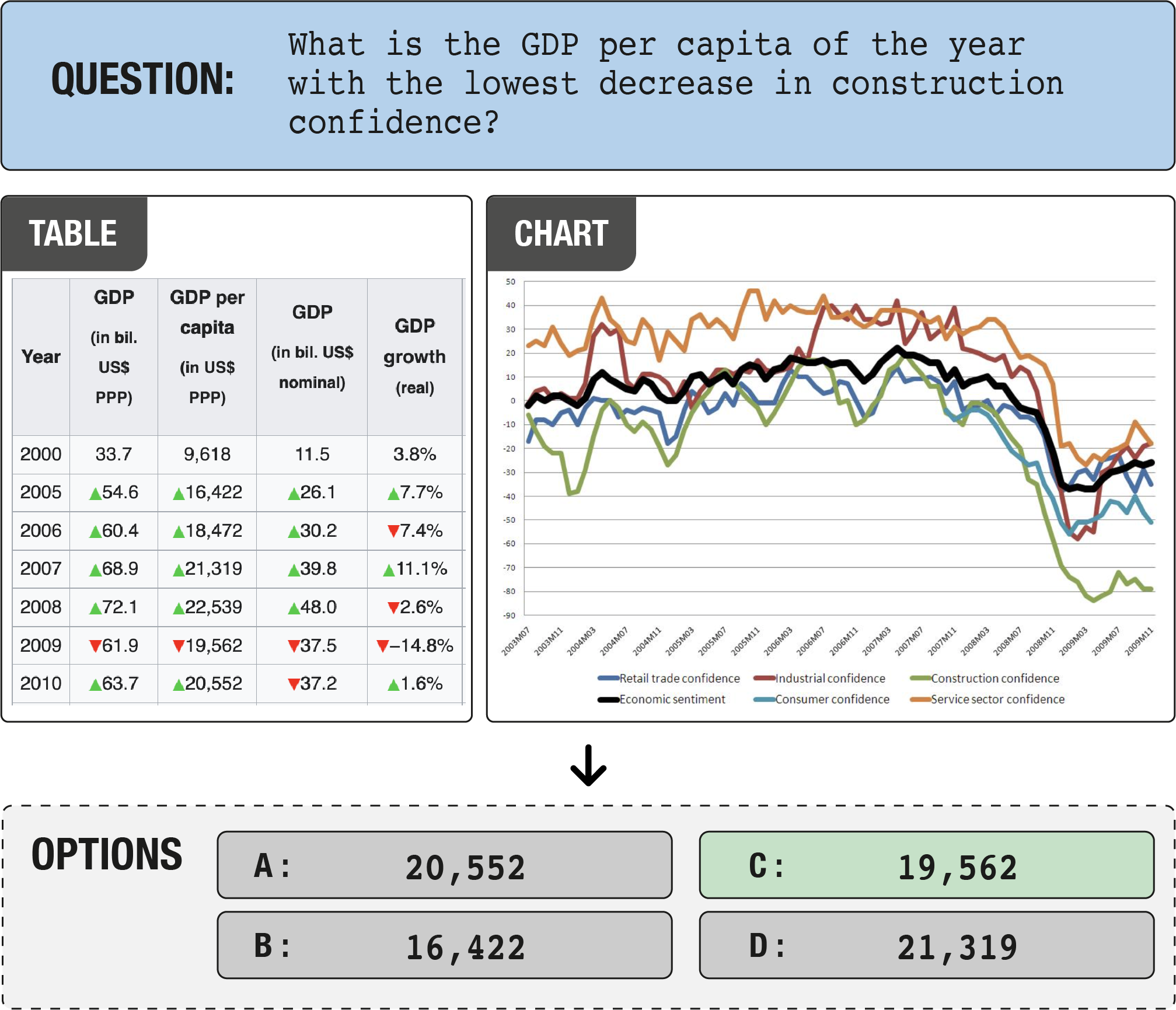}
    \caption{An example from \DATASET{} illustrating a question whose answer relies on the information presented in the accompanying table and chart.} 
    \label{fig:main}
\end{figure}

\begin{figure*}[ht]
    \centering
    \includegraphics[width=\linewidth]{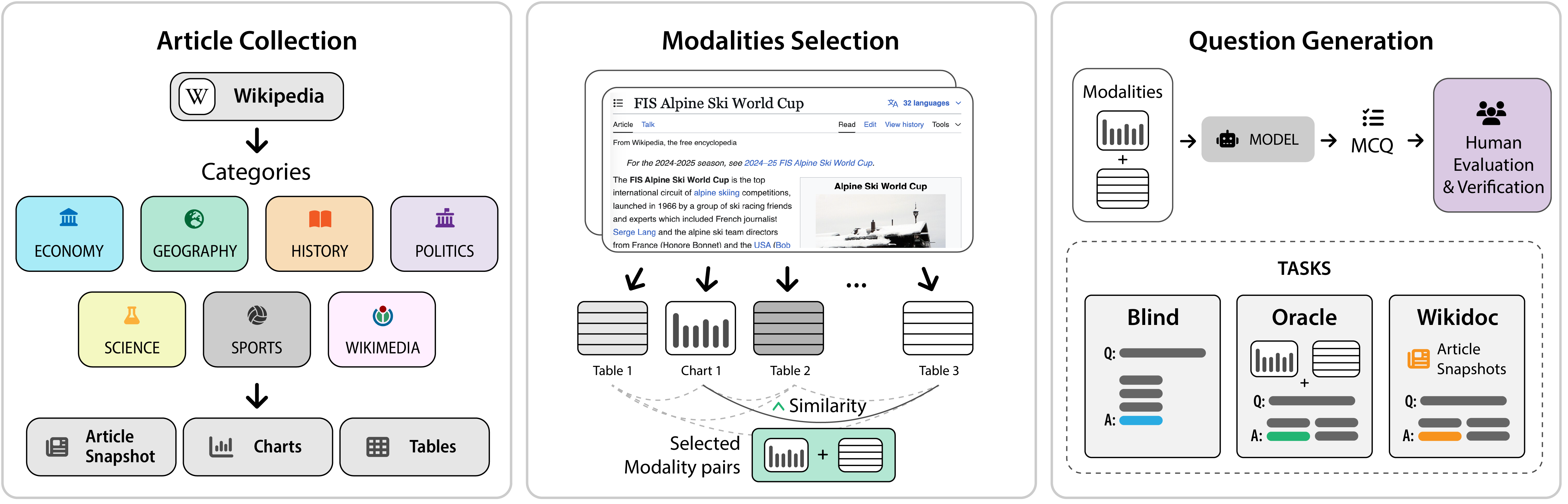}
    \caption{\DATASET{} Creation Pipeline:
(1) We collect Wikipedia articles that contain tables and charts.
(2) For each article, we identify table-chart pairs that exhibit semantic similarity.
(3) We employ GPT-4-turbo to generate multiple-choice questions (MCQs) based on each table-chart pair.
(4) Human annotators assess and validate the quality of the generated questions to ensure accuracy and relevance.}
    \label{fig:process}
\end{figure*}

A notable challenge in DU arises from the prevalence of documents containing large tables and charts, which can be difficult for humans to process and analyze. 
A question-answering (QA) system would help humans get insights from documents with such interjections easily.
Over the past few years, several Visual Question Answering (VQA) benchmarks have been developed to assess the DU capabilities of vision-language large models (VLLMs) across various aspects, including handling tables, charts, and document layouts. However, most existing benchmarks primarily focus on single-page documents.

Another key challenge in automatic DU is the ability to answer complex questions that require integrating information across multiple sections of a document and reasoning over different modalities. Current benchmarks largely lack multi-hop questions where models must synthesize information from multiple modalities—such as text, tables, and charts—to derive correct answers \citep{ma2024mmlongbench, van2023document}.
Furthermore, existing datasets lack controlled evaluation settings that isolate the specific modalities and types of information required to answer a question, making it difficult to conduct fine-grained analyses of VLLMs' limitations in DU tasks.

To bridge this gap, we introduce \DATASET, a benchmark dataset constructed using Wikipedia as the primary source of documents. 
\DATASET{} focuses on long, digital-only documents that contain multiple large tables and charts. The questions in our dataset are specifically designed to require multimodal reasoning over these structured elements, covering a broad spectrum of topics.
The dataset comprises 1,000 multiple-choice questions (MCQs) derived from $\sim$4k Wikipedia pages, with each document averaging 24.18 pages and $1815.01 \pm 2825.16$ textual tokens.
The questions span diverse domains, including \textit{Economy, Geography, History, Politics, Science, Sport}, and \textit{Wikimedia}. To ensure the need for multimodal reasoning, questions are structured to require information from either two tables (table-table), two charts (chart-chart),\footnote{We define a chart as a graphical representation of data, including: (a) data charts such as diagrams or graphs that organize and display numerical or qualitative information; (b) maps enhanced with additional data; and (c) other domain-specific constructs, such as chord charts or record charts.} or a combination of one table and one chart (table-chart).

The dataset construction follows a systematic pipeline: (1) We collect approximately 7,200 Wikipedia pages containing at least three tables and one chart; (2) We identify highly similar table-table, chart-chart, or table-chart pairs to ensure meaningful cross-modal reasoning; (3) We employ GPT-4-turbo to generate MCQs based on predefined criteria; and (4) A rigorous human curation process is conducted to refine and validate 1,000 fully curated questions.

We conduct an extensive evaluation on \DATASET{} using four open-source and eight closed-source state-of-the-art VLLMs across three different experimental settings, where we vary the level of contextual information provided to the models. The results, summarized in Table \ref{tab:model_accuracy}, reveal that while closed-source models perform relatively well ($\sim$70\%) when provided with the exact relevant information, they struggle significantly when required to retrieve relevant context from long documents before answering the questions. Notably, GPT-4-o is the only model to exceed 50\% accuracy in such a setting, while other closed-source models exhibit near-random performance. Open-source models perform even worse, with the highest accuracy reaching only 27\% when exact information is provided as input. These findings underscore the persistent challenge of long-context multimodal document understanding for VLLMs.

\section{Dataset Construction}

\begin{figure*}[ht]
    \centering
    \includegraphics[width=\linewidth]{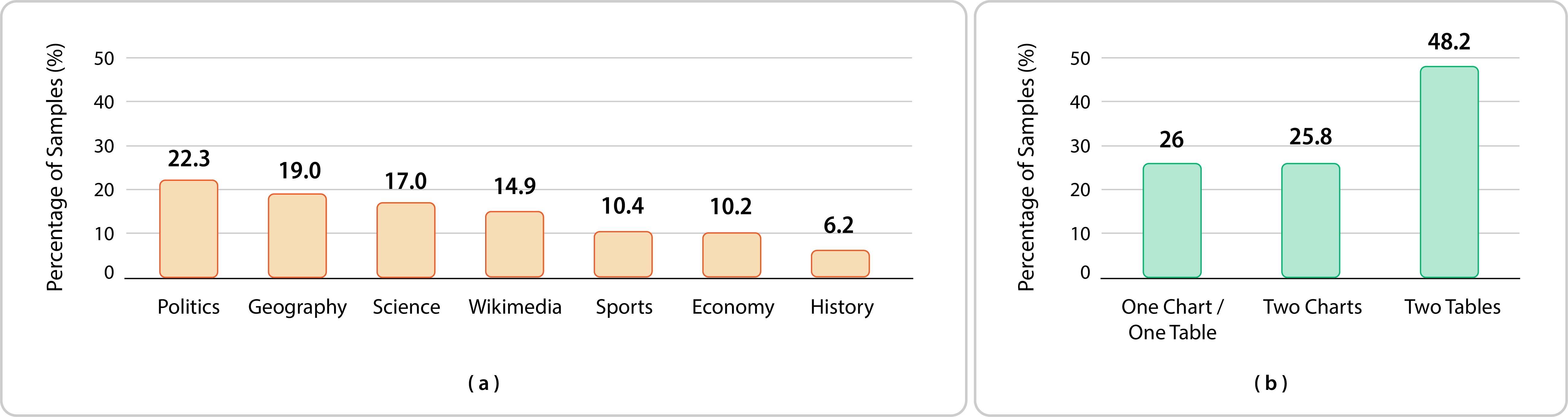}
    \caption{(a) Distribution of question-answer pairs across seven topics. (b) Distribution of question-answer pairs by modality type.}
    \label{fig:data_distribution}
\end{figure*}

This section outlines the pipeline employed for collecting Wikipedia documents, extracting and selecting their associated modalities (\ie, tables and charts), and generating the multiple-choice question (MCQ) samples that constitute \DATASET. An overview of the \DATASET{} creation pipeline is illustrated in \ref{fig:process}, with a detailed explanation provided in the Appendix \ref{apdx:data}.

\subsection{Document collection}
\label{sec:doc-collection}

To construct the dataset, we used the WTabHTML\footnote{Available at \url{https://github.com/phucty/wtabhtml}} project’s preprocessed English Wikipedia dumps from March 2022, initially comprising over 4 million entries. We filtered out articles with fewer than three tables to eliminate small, less relevant tables, narrowing the set to 392,223 entries.

To ensure multimodality, we downloaded over a million images from these articles and filtered out non-chart images using a fine-tuned Vision Transformer (ViT) model,\footnote{\url{https://huggingface.co/facebook/dinov2-base-imagenet1k-1-layer}} retaining relevant formats like PNG and JPEG. Articles with at least one valid chart were further filtered, reducing the set to 15,164 entries. To promote diversity, we categorized each document using Wikipedia’s \textit{``Instance of''} property and grouped similar categories into a custom taxonomy of seven main categories: \textit{Economy, Geography, History, Politics, Science, Sport,} and \textit{Wikimedia}. This final step reduced the dataset to 7,258 documents, ensuring broad coverage across various topics. Table ~\ref{tab:doc_stats} shows the document statistics for each category.

\subsection{Modalities selection}
\label{sec:mod-selection}

Wikipedia documents often contain multiple tables and charts. If questions are generated from randomly selected tables and charts, the resulting question set may lack diversity. Furthermore, if the selected tables and/or charts are not semantically relevant, generating meaningful and challenging questions becomes infeasible. To address these issues, we focus on generating questions that involve structured modality pairs: two tables (\textit{table-table}), one table and one chart (\textit{table-chart}), or two charts (\textit{chart-chart}). Figures \ref{fig:main} and \ref{fig:examples} show examples for each of these question types. 

To avoid irrelevant pairings, we selected pairs based on the textual similarity of the descriptions of each pair. Since most tables lacked captions, we used the \texttt{Llama-3-8B-Instruct}\footnote{\href{https://huggingface.co/meta-llama/Meta-Llama-3-8B-Instruct}{https://huggingface.co/meta-llama/Meta-Llama-3-8B-Instruct}} language model to generate descriptions from their raw HTML. For images identified as potential charts, we employed the vision-language model \texttt{GPT-4-turbo} to confirm whether they were charts and extract key information in fewer than 200 words. This approach ensured meaningful modality descriptions, with full prompt details provided in Appendix \ref{llama3-prompt}.

\begin{figure}[ht]
    \centering
    \includegraphics[width=0.9\linewidth]{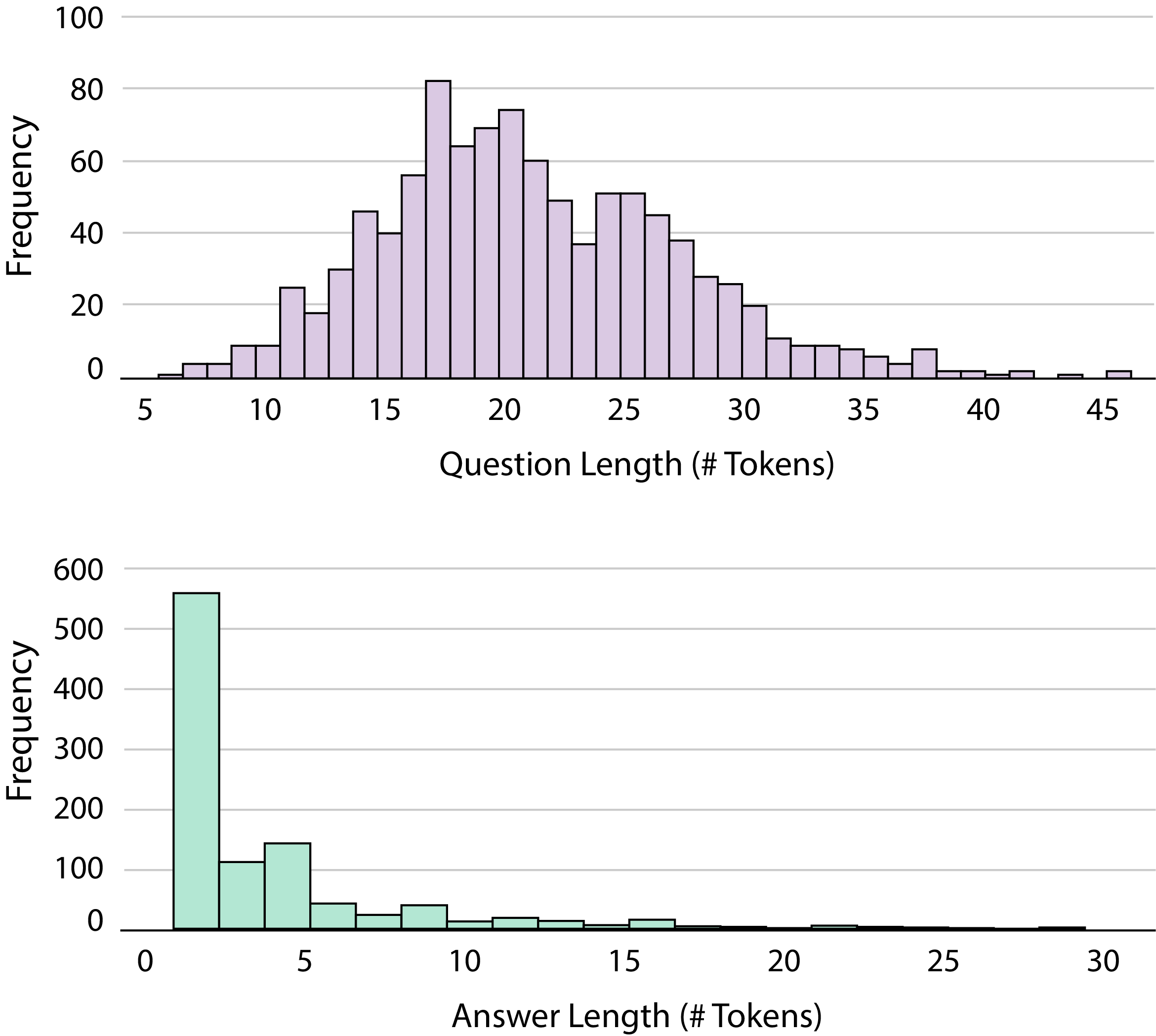}
    \caption{Distribution of questions and answers lengths.} 
    \label{fig:histigrams}
\end{figure}

Once we generated textual descriptions for both HTML tables and images, we calculated similarity scores for each possible modality pair (\textit{table-table}, \textit{table-chart}, and \textit{chart-chart}) within a document. We used the \textit{cross-encoder} model \texttt{BAAI/bge-reranker-v2-m3}~\cite{chen2024bge}, available on HuggingFace, which directly outputs a similarity score for two inputs instead of generating embeddings. Although slower than embedding models, this reranker model provides more accurate results. Since each document contains a limited number of modalities, this approach balances speed and accuracy effectively. Only images identified as charts by GPT-4-turbo were considered for this process. Appendix~\ref{chart_examples} provides more information on what we consider as charts.

\subsection{Question generation}
\label{sec:question-gen}

To ensure meaningful and challenging questions, we filtered modality pairs (\textit{table-table}, \textit{table-chart}, and \textit{chart-chart}) based on similarity scores. We calculated the macro mean similarity score for each pair type across topics and retained pairs with scores between the macro mean and 0.9. Tables with fewer than 512 characters were also excluded for their limited information content. We aimed for a balanced distribution of modality types and topics by selecting an equal number of pairs per type and topic, capping the number of pairs from the same document. Three types of multiple-choice questions were generated: two using individual modalities and one combining both. Each question had four options, one correct answer, and an explanation. Ultimately, 3,528 question-answer pairs were generated using GPT-4-turbo.

\paragraph{Quality Control} Through manual inspection of the generated question-answer pairs, we observed that some of the generated questions were invalid.
Despite efforts to select related modalities, there were cases where the information between the two modalities did not overlap (\eg, tables with differing date ranges or charts representing unrelated regions). To reduce the number of invalid questions reaching the annotation phase, we utilized a state-of-the-art vision-language model, \texttt{OpenGVLab/InternVL2-Llama3-76B}, available on HuggingFace\footnote{Available at \url{https://huggingface.co/OpenGVLab/InternVL2-Llama3-76B}}.

The model was provided with both modalities, and we prompted it to determine whether sufficient information was present in the charts and/or HTML tables to answer the given question (see the full prompt in Appendix~\ref{app:internvl2}). If the model’s response was ``yes,'' we followed up with the question:
Is this answer correct: \texttt{gpt4\_full\_answer}? Answer with ``yes'' or ``no.''
Here, \texttt{gpt4\_full\_answer} refers to the answer originally suggested by GPT-4 during the generation process. Only pairs passing this two-step evaluation were retained for further processing.

\subsection{Human curation}
\label{sec:human-eval}

Out of the 3,528 generated candidates, we selected 2,001 question-answer pairs for annotation. This included 938 pairs positively evaluated by the InternVL2 model and 1,063 pairs randomly sampled from the remaining dataset. Three Master’s students in Computer Science annotated all the selected pairs.
The annotation process consisted of two steps: 
\begin{enumerate}
    \item \textit{Validity Check}:  Annotators first determined if a question could only be answered by integrating information from both provided modalities. This ensured there was no informational overlap between modalities and that both were essential context for the question.
    \item \textit{Answer Assessment}:  For question-answer pairs deemed valid, annotators assessed the correctness of the provided answers, labeling each pair as ``Correct'', ``Wrong'', or ``Small Edit''. The ``Small Edit'' label was used for cases where the question could be retained after minor revisions. During this step, annotators also verified that incorrect answer options were plausible and contextually grounded.
\end{enumerate}
Using majority voting, 595 questions were labeled as Correct.
Invalid pairs were revised for issues like overly detailed questions or multiple correct answers. Pairs labeled ``Small Edit'' were refined, resulting in 405 additional corrected pairs being added to the final dataset. Figure \ref{fig:annotator} shows the interface of our annotation tool.

\subsection{The \DATASET{} benchmark}
Combining questions labeled as \textit{Correct} with the revised questions, the final dataset comprises 1000 question-answer pairs derived from 526 unique Wikipedia documents. The distribution of question-answer pairs across topics is relatively balanced, reflecting the natural topical distribution of Wikipedia, as shown in Figure~\ref{fig:data_distribution}.

Approximately 515 of the 1000 pairs were validated as \textit{Correct} by our AI evaluator (\texttt{InternVL2-Llama3-76B} model), underscoring the value of sampling from initially rejected questions to enhance dataset diversity. The distribution of question-answer pairs by modality type is illustrated in Figure~\ref{fig:data_distribution}. Notably, nearly half of the pairs involve reasoning across two tables. Figure~\ref{fig:histigrams} indicates the distribution of questions' and answers' lengths.
\section{Evaluation}
\label{evaluation}

\paragraph{Evaluation Setup} To rigorously assess the performance of state-of-the-art Vision-Language Learning Models (VLLMs), we design three distinct evaluation setups that vary the amount of contextual information supplied to the model. These setups are defined as follows:
(1) \texttt{blind}: In this scenario, no contextual information is provided to the model. The model is tasked with answering the question based solely on its internal knowledge or reasoning capabilities.
(2) \texttt{oracle}: Here, the model is supplied with the necessary visual or tabular data, such as charts or tables, that are essential for answering the question. This setup isolates the model's ability to interpret and reason with the provided structured data
(3) \texttt{wikidoc}: In this case, the model is given snapshots of the Wikipedia page from which the question was derived. This setup evaluates the model's capacity to process and utilize textual information from a comprehensive and unstructured source. Due to thecomputational-heavy nature of this setup, we only use it for closed-source models.

\paragraph{Models} 
We conduct an evaluation of the benchmark using state-of-the-art vision large language models (VLLMs), encompassing both open-source and closed-source models across various scales.
For open-source models, we include the Qwen family of models \citep{Yang2024Qwen2TR} (\textit{Qwen2-VL-7B-Instruct} and \textit{Qwen2-VL-72B-Instruct}), OpenGVLab's InternVL2 \citep{chen2023internvl} series (\textit{InternVL2.5-1B}, \textit{InternVL2.5-5B}, \textit{InternVL2.5-26B}, \textit{InternVL2.5-78B}), and Meta's Llama \citep{dubey2024llama} series (\textit{Llama-3.2-11B-Vision-Instruct}).
For closed-source models, the benchmark evaluation is conducted on \textit{GPT-4o} \citep{achiam2023gpt}, \textit{Gemini-1.5-Flash}, \textit{Gemini-1.5-Pro} \citep{team2024gemini}, and \textit{Claude3.5-Sonnet} \citep{claude3.5}. To evaluate open-source models, we use the vLLM library,\footnote{https://docs.vllm.ai/en/latest/} on a machine with 8 NVIDIA A100 GPUs (40GB memory).

\paragraph{Evaluation Metrics}
For each evaluation setting, we provide the model with a set of questions and their corresponding contextual information, prompting it to generate the correct answer choice (\ie, A, B, C, or D). We then assess model performance by measuring \textit{accuracy} and comparing results across different models. Appendix \ref{ap:answer_generation_prompts} provides details regarding the prompt design and usage.

\section{Analysis}

\begin{table}
\centering
\resizebox{\columnwidth}{!}{
\begin{tabular}{lccc}
\toprule
\textbf{Model} & \textbf{Blind} & \textbf{Oracle} & \textbf{Wikidoc} \\
\midrule
GPT-4o   & 33.46 & 71.42 & 55.24 \\
Gemini-2.0-pro & 22.67 & 69.53 & 23.47 \\
Gemini-2.0-flash & 23.27 & 67.52 & 24.47 \\
Claude3.5-Sonnet & 11.28 & 70.82 & 35.56 \\
\midrule
InternVL2.5-1B-MPO  & 19.88 & 23.17 &- \\
InternVL2.5-5B-MPO  & 22.17 & 27.87 &- \\
InternVL2.5-26B-MPO & 11.48 & 26.37 &- \\
InternVL2.5-78B-MPO & 03.09 & 27.37 &- \\
InternVL2.5-78B     & 03.29 & 27.67 &- \\
Qwen2.5-VL-7B-Instruct & 12.68 & 22.87 & - \\
Qwen2.5-VL-72B-Instruct & 0.39 & 23.17 & - \\
Llama-3.2-11B-Vision-Instruct & 10.68 & 14.08 & - \\
\midrule
Human Experts & - & 87.50 & - \\
\bottomrule
\end{tabular}
}
\caption{Models' performance (\textbf{accuracy \%}) under three different evaluation settings. Random baseline is 25\%. 
} 
\label{tab:model_accuracy}
\end{table}

\begin{table*}[h!]
\centering
\resizebox{\textwidth}{!}{
\begin{tabular}{lccccccc}
\toprule
\textbf{Model} & \textbf{History} & \textbf{Politics} & \textbf{Geography} & \textbf{Sports} & \textbf{Science} & \textbf{Economy} & \textbf{Wikimedia} \\
\midrule
GPT-4o                      & 74.12 & 68.61 & \textbf{76.32} & \textbf{75.96} & \textbf{72.94} & 58.25 & 72.48 \\
Gemini-2.0-flash            & 74.19 & 65.02 & 72.11 & 70.19 & 65.88 & 56.31 & 70.47 \\
Gemini-2.0-pro              & \textbf{75.81} & 71.75 & 72.11 & 69.23 & 69.41 & 52.43 & 72.48 \\
Claude3.5-Sonnet             & 67.74 & \textbf{76.68} & 69.47 & 71.15 & 68.24 & \textbf{61.17} & \textbf{74.50} \\
\midrule
InternVL2.5-1B-MPO               & 30.65 & 22.87 & 18.95 & 25.00 & 25.29 &  15.53 & 27.52 \\
InternVL2.5-5B-MPO             & 32.26 & 30.49 & 28.42 & 21.15 & 27.65 & 25.24 & 28.19 \\
InternVL2.5-26B-MPO               & 32.26 & 26.01 & 25.79 & 24.04 & 24.12 & 23.30 & 31.54 \\
InternVL2.5-78B-MPO             & 33.87 & 22.42 & 27.89 & 27.88 & 24.71 & 32.04 & 30.87\\
InternVL2.5-78B            & 35.48 & 24.66 & 25.79 & 31.73 & 24.12 & 33.01 & 28.86 \\
Qwen2.5-VL-7B-Instruct        & 27.42 & 22.87 & 17.37 & 23.08 & 24.12 & 21.36 & 27.52 \\
Qwen2.5-VL-72B-Instruct       & 30.65 & 22.87 & 18.95 & 25.00 & 25.29 & 15.53 & 27.52 \\
Llama-3.2-11B-Vision-Instruct & 06.45 & 14.35 & 16.84 & 14.42 & 17.06 & 11.65 & 11.41  \\
\bottomrule
\end{tabular}
}
\caption{Models' performance (\textbf{accuracy \%}) across various topics in the \texttt{oracle} evaluation setting.} 
\label{tab:topic_accuracy}
\end{table*}
 
Table~\ref{tab:model_accuracy} presents the performance of the evaluated models on \DATASET.
In the \texttt{blind} setting, where models lack access to contextual information, all models perform below random chance, with the exception of GPT-4o, which achieves a slightly higher accuracy of 33\%. This result is expected, as questions in \DATASET{} require contextual information to be answered correctly, and without such context, models are unable to make informed predictions.

In the \texttt{oracle} setting, where models receive direct access to relevant context, proprietary models perform significantly better. GPT-4o achieves the highest accuracy, with Claude and Gemini models closely following. In contrast, open-source models exhibit poor performance, performing at or near random levels.

For the \texttt{wikidoc} setting, we exclude closed-source models due to their limited context length, which prevents them from processing the full Wikipedia snapshots required for answering questions. Among open-source models, only GPT-4o surpasses 50\% accuracy. This observation suggests that while these models perform well when provided with explicitly relevant information, they struggle when required to process long-context inputs. The challenge arises from their need to first locate relevant information within extensive text before formulating an answer, highlighting a key limitation in handling extended vision-context scenarios.

In our human evaluation, annotators achieved an accuracy of 87\% in the \texttt{oracle} setting, revealing a substantial performance gap of approximately 17\% between current VLLMs and human performance. This discrepancy underscores the challenges that LVLMs face in document understanding and highlights the necessity of our benchmark for advancing research in this area. 

\paragraph{Question Type Analysis}
\begin{table}[]
\centering
\resizebox{\columnwidth}{!}{
\begin{tabular}{lccc}
\toprule
\textbf{Model} & \textbf{2 Charts} & \textbf{2 Tables} & \textbf{1 Chart/1 Table} \\
\midrule
GPT-4o   & 71.31 & 71.63 & 71.15 \\
Gemini-2.0-flash & 53.48 & 74.12 & 69.23 \\
Gemini-2.0-pro & 54.65 & 77.43 & 69.61 \\
Claude3.5-Sonnet & 66.66 & 73.29 & 70.38 \\
\midrule
InternVL2.5-1B-MPO & 20.93 & 24.43 & 23.07 \\
InternVL2.5-8B-MPO & 25.19 & 31.26 & 24.23 \\
InternVL2.5-26B-MPO & 24.03 & 29.19 & 23.46 \\
InternVL2.5-78B-MPO & 24.41 & 29.39 & 26.53 \\
InternVL2.5-78B & 24.41 & 30.02 & 26.53 \\
Qwen2.5-VL-7B-Instruct & 18.99 & 24.63 & 23.46 \\
Qwen2.5-VL-72B-Instruct & 22.48 & 24.22 & 21.92 \\
Llama-3.2-11B-Vision-Instruct & 28.29 & 02.27 & 21.92 \\
\bottomrule
\end{tabular}
}
\caption{Models' performance (\textbf{accuracy \%}) across three question types in the \texttt{oracle} setting.} 
\label{tab:qtype_accuracy}
\end{table}
 
Table \ref{tab:qtype_accuracy} presents a breakdown of model performance across three question types: (i) questions involving two charts, (ii) questions involving two tables, and (iii) questions involving a combination of one chart and one table.

For questions that require interpreting two charts, GPT-4o demonstrates the highest performance, outperforming Claude and Gemini models by 5\% and 17\%, respectively. This result suggests that GPT-4o is particularly effective at processing and reasoning over chart-based data compared to other models.

In contrast, for questions involving two tables, GPT-4o exhibits a relatively consistent performance across all question types. However, Gemini-2.0-pro achieves the highest accuracy in this category, indicating its superior capability in handling tabular data.

Finally, for questions that involve both a chart and a table, all proprietary models achieve similar performance levels. Given that closed-source models perform at a level indistinguishable from random chance, further fine-grained analysis is not meaningful in this context.

\paragraph{Topic Analysis}

Table \ref{tab:topic_accuracy} presents a breakdown of models performance across different question types in the \texttt{oracle} setting. The results indicate that models perform relatively consistently across various topics, with the exception of the Economy topic, where all models exhibit slightly lower performance. A potential explanation for this discrepancy is that Economy-related questions frequently involve bar and line charts, necessitating both chart interpretation and comparative analysis. These tasks may pose greater challenges for the models, leading to a decline in performance on these questions.


\begin{table*}[ht]
\centering
\resizebox{0.96\textwidth}{!}{
\setlength{\tabcolsep}{4pt} 
\renewcommand{\arraystretch}{1.1} 
\small 
\begin{tabularx}{\textwidth}{@{}l l r r r r r r@{}}
\toprule
\textbf{Topic} & \textbf{Subtopic} & \textbf{Docs} & \textbf{Images} & \textbf{Charts} & \textbf{Tables} & \textbf{Table Rows} & \textbf{Tokens} \\
\midrule
Economy   & Budget         & 3   & 1.33 $\pm$ 0.58  & 0.67 $\pm$ 0.58 & 3.00 $\pm$ 0.00 & 21.78 $\pm$ 12.47 & 1626.33 $\pm$ 936.56 \\
          & GDP            & 60  & 3.13 $\pm$ 2.71  & 2.27 $\pm$ 2.50 & 5.37 $\pm$ 4.19 & 15.63 $\pm$ 13.12 & 6465.23 $\pm$ 3028.17 \\
          & Reform         & 1   & 4.00             & 0.00           & 5.00           & 7.00 $\pm$ 1.67   & 2901.00 \\
          & Stock market   & 15  & 1.20 $\pm$ 0.41  & 1.00 $\pm$ 0.53 & 3.67 $\pm$ 1.07 & 29.29 $\pm$ 35.65 & 994.40 $\pm$ 907.67 \\
          & Tax            & 1   & 4.00             & 4.00           & 5.00           & 5.20 $\pm$ 2.71   & 11179.00 \\
          \cmidrule{2-8}
          & Total / Avg          & 80 & 2.73 $\pm$ 2.47   & 1.96 $\pm$ 2.25 & 4.95 $\pm$ 3.73 & 17.43 $\pm$ 18.61 & 5272.36 $\pm$ 3545.17\\    
\midrule
Geography & City           & 509 & 1.28 $\pm$ 0.61  & 0.35 $\pm$ 0.58 & 4.52 $\pm$ 3.42 & 10.21 $\pm$ 10.36 & 4307.26 $\pm$ 4144.97 \\
          & Country        & 87  & 2.89 $\pm$ 1.98  & 1.36 $\pm$ 1.44 & 6.36 $\pm$ 5.86 & 13.72 $\pm$ 21.72 & 8364.64 $\pm$ 5031.71 \\
          & Region         & 460 & 2.04 $\pm$ 2.56  & 0.92 $\pm$ 2.12 & 5.39 $\pm$ 3.13 & 11.17 $\pm$ 12.56 & 4384.26 $\pm$ 4183.70 \\
          & Transport      & 242 & 1.16 $\pm$ 0.46  & 0.36 $\pm$ 0.50 & 3.94 $\pm$ 1.51 & 10.55 $\pm$ 13.41 & 2534.54 $\pm$ 2539.62 \\
          \cmidrule{2-8}
          & Total / Avg          & 1298 & 1.63 $\pm$ 1.74  &0.62 $\pm$ 1.42 & 4.84 $\pm$ 3.35 & 10.95 $\pm$ 13.09 & 4275.99 $\pm$ 4186.91\\    
\midrule
History   & Battle         & 19  & 1.89 $\pm$ 1.37  & 0.58 $\pm$ 1.12 & 3.84 $\pm$ 2.30 & 11.67 $\pm$ 10.22 & 8128.32 $\pm$ 4793.27 \\
          & Dynasty        & 8   & 2.12 $\pm$ 1.46  & 0.62 $\pm$ 0.74 & 6.00 $\pm$ 3.81 & 11.17 $\pm$ 15.79 & 2532.62 $\pm$ 2023.55 \\
          & Other          & 38  & 2.29 $\pm$ 1.64  & 0.89 $\pm$ 1.06 & 6.32 $\pm$ 5.28 & 12.05 $\pm$ 13.59 & 6164.74 $\pm$ 3820.45 \\
          \cmidrule{2-8}
          & Total / Avg          & 65 & 2.15 $\pm$ 1.52   & 0.77 $\pm$ 1.03 & 5.55 $\pm$ 4.57 & 11.86 $\pm$ 13.31 & 6291.68 $\pm$ 4299.70\\    
\midrule
Politics  & \makecell{Composition of parlia-\\ment, government} & 1026 & 1.13 $\pm$ 0.63 & 0.24 $\pm$ 0.48 & 14.77 $\pm$ 14.80 & 8.35 $\pm$ 10.87 & 594.59 $\pm$ 1109.90 \\
          & Election results & 1382 & 1.45 $\pm$ 1.64 & 0.72 $\pm$ 0.90 & 14.58 $\pm$ 14.81 & 9.56 $\pm$ 14.14 & 1233.96 $\pm$ 1900.37 \\
          & Foreign relations & 9 & 1.22 $\pm$ 0.67 & 0.67 $\pm$ 0.71 & 5.89 $\pm$ 1.37 & 26.62 $\pm$ 15.67 & 2969.89 $\pm$ 1596.50 \\
          \cmidrule{2-8}
          & Total / Avg          & 2420 & 1.32 $\pm$ 1.31   & 0.52 $\pm$ 0.79 & 14.63 $\pm$ 14.79 & 9.07 $\pm$ 12.88 & 969.34 $\pm$ 1647.15\\    
\midrule
Science   & Astronomy      & 340 & 3.28 $\pm$ 2.41  & 0.25 $\pm$ 0.63 & 3.71 $\pm$ 0.97 & 8.31 $\pm$ 4.87   & 513.51 $\pm$ 457.57 \\
          & Biology        & 73  & 2.64 $\pm$ 1.64  & 1.11 $\pm$ 1.06 & 4.00 $\pm$ 1.57 & 10.03 $\pm$ 7.85  & 988.05 $\pm$ 539.53 \\
          & Chemistry      & 82  & 1.50 $\pm$ 0.95  & 0.67 $\pm$ 0.82 & 3.91 $\pm$ 1.06 & 6.51 $\pm$ 4.48   & 5215.02 $\pm$ 2306.80 \\
          & Demography     & 248 & 2.63 $\pm$ 2.51  & 1.77 $\pm$ 1.59 & 9.35 $\pm$ 7.36 & 16.37 $\pm$ 23.56 & 2695.69 $\pm$ 2716.78 \\
          \cmidrule{2-8}
          & Total / Avg   & 743 & 2.81 $\pm$ 2.33   & 0.89 $\pm$ 1.29 & 5.64 $\pm$ 5.07 & 12.75 $\pm$ 18.34 & 1807.38 $\pm$ 2356.32\\    
\midrule
Sport     & Events         & 319 & 1.11 $\pm$ 0.35  & 0.51 $\pm$ 0.57 & 10.07 $\pm$ 11.64 & 11.08 $\pm$ 17.66 & 1616.35 $\pm$ 1844.19 \\
          & Results        & 52  & 1.21 $\pm$ 0.67  & 0.62 $\pm$ 0.60 & 10.23 $\pm$ 9.49 & 10.62 $\pm$ 22.21 & 1364.02 $\pm$ 2596.41 \\
          & Teams          & 1264 & 1.11 $\pm$ 0.36  & 0.24 $\pm$ 0.44 & 10.51 $\pm$ 7.60 & 10.50 $\pm$ 12.10 & 1217.02 $\pm$ 1760.18 \\
          \cmidrule{2-8}
          & Total / Avg          & 1637 & 1.11 $\pm$ 0.37   & 0.31 $\pm$ 0.49 & 10.42 $\pm$ 8.60 & 10.61 $\pm$ 13.73 & 1299.26 $\pm$ 1815.95\\    
\midrule
Wikimedia     & Article         & 973 & 3.32 $\pm$ 8.74  & 0.87 $\pm$ 2.47 & 12.61 $\pm$ 12.46 & 16.17 $\pm$ 32.29 & 920.20 $\pm$ 1885.78 \\
          & Information        & 9  & 1.22 $\pm$ 0.44  & 0.33 $\pm$ 0.50 & 8.89 $\pm$ 7.23 & 18.49 $\pm$ 21.94 & 571.78 $\pm$ 582.64 \\
          & Person          & 4 & 2.00 $\pm$ 1.41  & 0.50 $\pm$ 0.58 & 16.50 $\pm$ 12.52 & 15.06 $\pm$ 35.96 & 508.25 $\pm$ 276.44 \\
          & Timeline          & 15 & 3.07 $\pm$ 4.15  & 0.67 $\pm$ 0.62 & 20.13 $\pm$ 18.03 & 19.60 $\pm$ 28.09 & 1278.53 $\pm$ 2995.44 \\
          & Overview          & 14 & 6.57 $\pm$ 8.89 & 5.14 $\pm$ 7.87 & 9.86 $\pm$ 14.63 & 12.16 $\pm$ 16.87 & 3935.00 $\pm$ 3021.31 \\
          \cmidrule{2-8}
          & Total / Avg       & 1015 & 3.34 $\pm$ 8.64   & 0.92 $\pm$ 2.63 & 12.66 $\pm$ 12.60 & 16.22 $\pm$ 32.05 & 962.36 $\pm$ 1948.86\\    
\midrule
\midrule
\textbf{Total / Avg} & \textbf{-} 
    & 7258 
    & 1.79 $\pm$ 3.59
    & 0.60 $\pm$ 1.37
    & 10.55 $\pm$ 11.48
    & 11.02 $\pm$ 18.25
    & 1815.01 $\pm$ 2825.16 \\
\bottomrule
\end{tabularx}
}
\caption{Document Statistics by Topic and Subtopic.}
\label{tab:doc_stats}
\end{table*}
\section{Related Work}

Visual Question Answering (VQA) is a crucial subtask of document understanding (DU), where the objective is to generate natural language answers to questions based on a given visual document.
Previous research has explored the DU capabilities of vision large language models (VLLMs) by introducing new datasets and benchmarks. Many of these datasets are designed to evaluate specific components, such as tables \citep{herzig2021open, chen2021ottqa} or charts \citep{chaudhry2020leaf, Methani_2020_WACV, masry-etal-2022-chartqa, kantharaj-etal-2022-opencqa, SlideVQA2023}, and are often constrained to single-page documents.

\begin{table*}[]
\centering
\resizebox{\textwidth}{!}{
\begin{tabular}{lcccccccc}
\toprule
\textbf{Benchmark} & Sources & Origin & \# Docs. & \# Questions & Cross-page & \# Tokens & Answer Type & Evidence Source \\

\midrule
MMLongBench-Doc & Multi  & Digital + scan & 135 & 1k & \checkmark &  21214.1 &  Abst + Ext. & T, L, F, Ch, M\\
DUDE & Multi & Digital + scan & 5k & 41k & \checkmark & 1,831.53 & Abst. + Ext. & T, L, F, Ch, M\\
MP-DocVQA & Industry docs & Mostly scans & 6k & 50k & \xmark & 2026.6 & Ext. & T, L, F, Ch\\
VisualMRC & Web pages & Digital & 10k & 30k & \xmark &  154.19 & Abst. & T, L, F, Ch\\
InfographicsVQA & Infographics & Digital & 5.4k & 30k & \xmark &  287.98& Abst + Ext. & T, L, F, Ch, M\\
TAT-DQA & Finance reports& Digital & 2.7k & 16k & \xmark & 576.99 & Abst. + Ext.& T, L\\
\midrule
\DATASET & Wikipedia & Digital & $\sim4k$ & 1k & \checkmark &  1815.01 &  Abst. + Ext.& T, L, F, Ch, M\\

\bottomrule
\end{tabular}
}
\caption{Comparison between \DATASET{} and existing VQA benchmarks. Evidence Sources are abbreviated as (T)able, (L)ist, (F)igure, (Ch)art, and M(ap). Answer types are Extractive (Ext.) and Abstractive (Abst.)}
\label{tab:dataset_comparision}
\end{table*}

While recent benchmarks have attempted to extend document VQA beyond single-page settings, they still face limitations in terms of cross-page reasoning, domain diversity, and question complexity. For example, MP-DocVQA~\cite{tito2023hierarchical}, an extension of DocVQA~\citep{mathew2021docvqa}, does not include cross-page questions. 
DUDE \citep{van2023document} introduces a small proportion of cross-page questions but is limited by its reliance on crowd-sourced annotations, which often result in less challenging and rigorous questions, many of which focus on document layout rather than deeper content understanding. 
Similarly, SlideVQA~\citep{tanaka2023slidevqa} incorporates cross-page questions but is tailored to slide decks, which typically contain lower information density compared to other document types.
Doc2SoarGraph \citep{zhu-etal-2024-doc2soargraph} includes a few multi-page queries but remains limited to PDFs as its data format. 

FinanceBench~\citep{islam2023financebench} addresses some of these challenges by including long-context documents and practical cross-page questions. However, its exclusive focus on financial reports and open-ended answer formats requires expert-level manual evaluation, restricting its applicability to broader domains. 
Similarly, CRAB \citep{romanou2023crab} focuses on question-answering based on events spanning multiple documents, however, its scope is limited to the domain of causal understanding. 
More recently, MMLongBench-Doc \citep{ma2024mmlongbench} was introduced, incorporating questions from diverse sources, with approximately one-third being cross-page questions. Despite this, the dataset is derived from only 130 documents, limiting its domain coverage and diversity.

A more detailed comparison of existing datasets is presented in Table \ref{tab:dataset_comparision}, highlighting the unique contributions of \DATASET{} in advancing research on multi-modal reasoning and document-based question answering.

\section{Conclusion}
In this paper, we introduce \DATASET{}, a multi-modal visual question-answering (VQA) benchmark designed to assess the long-context document understanding (DU) capabilities of vision-language large models (VLLMs). Our benchmark consists of 1,000 multiple-choice questions (MCQs) that necessitate complex, multi-hop reasoning over visual data, including charts and tables.

We conducted an extensive evaluation of both closed-source and open-source models. Our results indicate that while closed-source models perform well when provided with precisely relevant information, their performance degrades significantly in settings where they must process and extract relevant details from long-context visual data. In contrast, state-of-the-art open-source models exhibit performance close to random, suggesting fundamental challenges in their reasoning capabilities.

These findings highlight that current VLLMs struggle with VQA tasks requiring the extraction and integration of dispersed information from extended contexts. We hope that \DATASET{} serves as a valuable resource for the research community in identifying and addressing the limitations of VLLMs in reasoning and long-context visual understanding.
\section{Limitations}

\DATASET{} is constructed using Wikipedia as the primary source of documents. Consequently, the generated questions are constrained to Wikipedia-style content, and the benchmark currently covers only seven topics.

A key characteristic of this benchmark is that it includes questions requiring information from multiple modalities. However, this introduces a limitation, as the dataset does not yet support more complex multi-hop reasoning. To address this, we plan to release the full dataset, including the filtered Wikipedia pages, along with the extracted charts and tables. This will enable future research to extend the dataset with more sophisticated question formulations.

Another limitation of this study is the long-context evaluation methodology. Specifically, our evaluation is conducted using image-based inputs (snapshots of Wikipedia pages), without incorporating the textual representations of these pages. Future work could enhance the evaluation by integrating text-based inputs to improve model performance and robustness.

\section*{Acknowledgment}

The authors would like to thank the anonymous reviewers for their insightful comments and constructive feedback. We are also grateful to the members of the LSIR and NLP laboratories at EPFL for their support and helpful input. Additionally, we thank Google for funding and supporting this project.

\bibliography{custom}

\appendix
\onecolumn
\section{Custom Wikipedia Category Taxonomy}
\label{sec:wikipedia_category_taxonomy}

\setlength\LTleft{0pt}
\setlength\LTright{0pt}

\begin{longtable}{|>{\raggedright\arraybackslash}p{2.5cm}|>{\raggedright\arraybackslash}p{3cm}|>{\raggedright\arraybackslash}p{9cm}|}
\hline
\textbf{Category} & \textbf{Subcategory} & \textbf{Instance of (\texttt{P31})} \\
\hline
\endfirsthead

\hline
\textbf{Category} & \textbf{Subcategory} & \textbf{Instance of (\texttt{P31})} \\
\hline
\endhead

\hline
\endfoot

\hline
\endlastfoot

\multirow{3}{2.5cm}{Politics}
    & Election results & *election*, legislative election, referendum \\ 
    & Composition of parliament, government & legislative district of *, electoral unit, constituency of the *, United States congressional district, federal electoral district of Canada, parliamentary constituency of *, political party, provincial electoral district of *, ward or electoral division of the United Kingdom, local government areas of * \\ 
    & Foreign relations & foreign policy, foreign relations \\ 
\hline
\multirow{3}{2.5cm}{Sport} 
    & Results & sports competition, Rugby World Cup qualification, Olympic medal table, championships, UEFA European Championship qualifying \\ 
    & Teams & association football club, association football team, college sports team, women's national association football team, association football team season, Olympic delegation, * team \\ 
    & Events & sport season, *Grand Prix, 24 Hours of Le Mans, Tour de France, Summer Olympic Games, Winter Olympic Games, Olympic sporting event, derby, nation at sport competition, multi-sport event, qualification event, qualification for the FIFA World Cup \\ 
\hline
\multirow{3}{2.5cm}{History} 
    & Battle & battle, war, world war \\ 
    & Dynasty & noble family, * dynasty \\ 
    & Other & aspect of history \\ 
\hline
\multirow{4}{2.5cm}{Science} 
    & Demography & demographics of country or region, ethnic group \\ 
    & Astronomy & * eclipse, asteroid, potentially hazardous asteroid, near-Earth object \\ 
    & Biology & gene, protein, group or class of transmembrane transport proteins, protein family associated with domain, protein family \\ 
    & Chemistry & chemical element, synthetic element \\ 
\hline
\multirow{4}{2.5cm}{Geography} 
    & City & city or town, big city, municipality of *, independent city, million city, city in the United States, human settlement, largest city, megacity, highly urbanized city \\ 
    & Country & country population, historical country, country \\ 
    & Region & geography of geographic location, region of *, regions of *, U.S. region, U.S. state, province of *, county, county of *, township of *, aspect in a geographic region, geographic region \\ 
    & Transport & rapid transit railway line, commercial traffic aerodrome, international airport, airport, railway station, railway line, airline, rapid transit, railway company, transport company \\ 
\hline
\multirow{5}{2.5cm}{Economy} 
    & GDP & national economy, regional economy \\ 
    & Stock market & stock market index \\ 
    & Budget & budget, military budget \\ 
    & Reform & economic reform \\ 
    & Tax & tax system \\
\hline
          & Article & Wikimedia list article \\ 
Wikimedia & Information & Wikimedia information list \\ 
          & Person & Wikimedia list of persons \\ 
          & Timeline & Wikimedia timeline \\ 
          & Overview & Wikipedia overview article \\ 
\hline
\caption{Categories and Subcategories with Examples} \label{tab:categories} \\
\end{longtable}
\twocolumn
\section{\DATASET creation}
\label{apdx:data}

\begin{figure*}
    \centering
    \includegraphics[width=\textwidth]{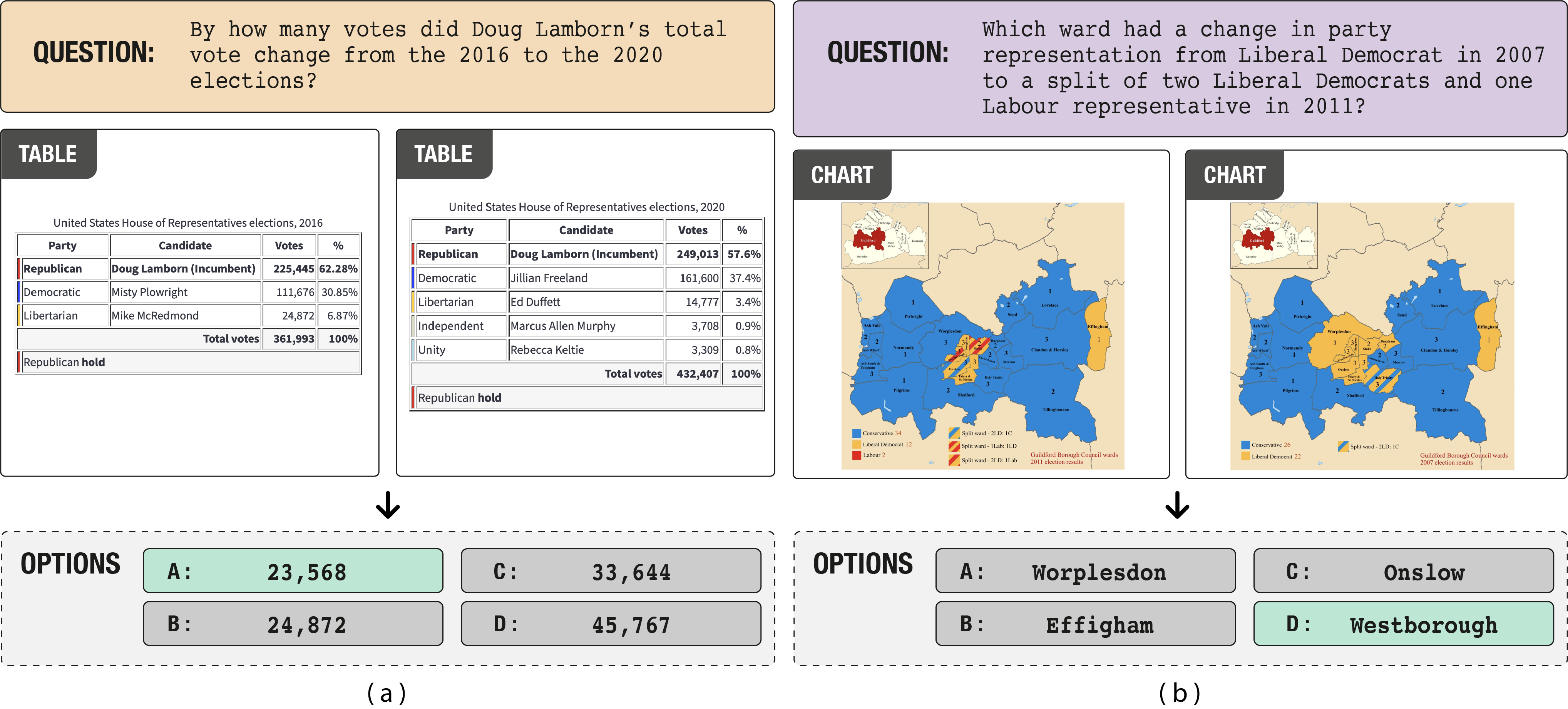}
    \caption{Examples from \DATASET{} illustrating (a) a question whose answer relies on the information presented in two tables and (b) a question whose answer relies on the information presented in two charts.} 
    \label{fig:examples}
\end{figure*}

\subsection{Wikipedia article collection}
\label{doc_selection}

To collect Wikipedia documents containing tables and charts, we leveraged the dataset curated by the WTabHTML project\footnote{Available at \url{https://github.com/phucty/wtabhtml}}. We downloaded preprocessed English-language dumps extracted from the 2022-03-01 Wikipedia dump. This dataset included 4,291,914 entries across 1,480,422 articles.

\begin{figure}[ht]
    \centering
    \includegraphics[width=\linewidth]{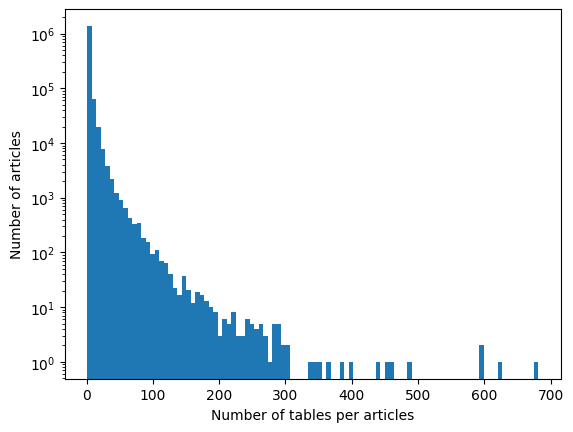}
    \caption{Distribution of tables in selected Wikipedia pages.}
    \label{fig:enter-label}
\end{figure}

Many articles contained very small tables (e.g., one-row sports results, as shown in Figure~\ref{fig:enter-label}), which were less useful for our purposes. To address this, we filtered out articles with fewer than three tables, resulting in a subset of 392,223 entries.

Next, we focused on articles containing charts to ensure a multimodal dataset. Images were downloaded from the 392,223 documents using the \texttt{pyWikiCommons} Python package, resulting in 1,041,062 images. Since most images in Wikipedia are natural images, flags, icons, or buttons rather than charts, we filtered them using a Vision Transformer (ViT) model trained with the DINOv2 method~\cite{oquab2023dinov2} and fine-tuned on ImageNet-1k categories. The model, available on HuggingFace as \texttt{facebook/dinov2-base-imagenet1k-1-layer}, classified images into relevant categories.
We restricted image formats to JPEG, JPG, and PNG, converting all SVG files to PNG beforehand. Images classified under \texttt{web site, website, internet site, site} ($\sim$90\%) or \texttt{oscilloscope, scope, cathode-ray oscilloscope, CRO} ($\sim$10\%, primarily line charts) were retained. Additional rules excluded irrelevant categories like flags and sports-related images based on filenames. Articles with at least one remaining chart were preserved, reducing the dataset to 15,164 entries.

\subsection{Promoting diversity}

To promote diversity in question generation, we sampled documents by category, aiming for a varied set of question-answer pairs. Each document was labeled with a category using the Wikipedia property \textit{Instance of (P31)}\footnote{See documentation at \url{https://www.wikidata.org/wiki/Property:P31}}, retrieved via the \texttt{pywikibot} Python package. Frequently occurring categories included politics and sports, covering documents like election results, parliamentary compositions, and sports team rosters.

To unify categories, we created a custom taxonomy grouping similar \textit{instance of} classes into subcategories (details in Appendix~\ref{sec:wikipedia_category_taxonomy}). For example, documents related to election results were grouped under the \textit{Election results} subcategory of \textit{Politics} using regular expressions (\texttt{*election*}). The final taxonomy included seven main categories: \textit{Economy}, \textit{Geography}, \textit{History}, \textit{Politics}, \textit{Science}, \textit{Sport}, and \textit{Wikimedia}. This step reduced the dataset to 7,258 documents, as shown in Table~\ref{tab:doc_stats}.

\paragraph{Balancing Categories}

The initial classification of charts using the ImageNet model, while efficient, resulted in a number of false positives. To improve chart identification, we utilized GPT-3.5-turbo to analyze image filenames and distinguish likely charts from non-charts (see Appendix~\ref{app-gpt-3-5}). In the most populated subtopics—such as Geography/City, Politics/Composition of Parliament or Government, Politics/Election Results, Science/Astronomy, and Sport/Teams—we excluded only documents that lacked any identified charts, as inferred by GPT-3.5. This refinement step reduced the dataset to 4,292 documents.

\paragraph{Downloading the Final Documents}

The final set of Wikipedia documents was downloaded in HTML format using the official API\footnote{\url{https://en.wikipedia.org/w/api.php}} in March and April 2024. Each HTML page was converted into a JPG image using the \texttt{imgkit} Python package\footnote{\url{https://github.com/jarrekk/imgkit}}. Due to large heights, images were split into segments of 768 pixels with a 32-pixel overlap using \texttt{pillow}\footnote{\url{https://github.com/python-pillow/Pillow}}. Note that some very long documents were truncated by \texttt{imgkit}, resulting in missing content. 
HTML tables were extracted using the WTabHTML extractor\footnote{\url{https://github.com/phucty/wtabhtml}}. 
Additionally, textual data was extracted for reference using the \texttt{wikipediaapi} Python package.

\subsection{Selection of modality pairs}
Wikipedia documents often contain multiple tables and charts. If questions are generated from randomly selected tables and charts, the resulting question set may lack diversity. Furthermore, if the selected tables and/or charts are not semantically relevant, generating meaningful and challenging questions becomes infeasible. To address these issues, we focus on generating questions that involve structured modality pairs: two tables (\textit{table-table}), one table and one chart (\textit{table-chart}), or two charts (\textit{chart-chart}).
To mitigate the selection of irrelevant pairs, we rely on the similarities in the textual descriptions between the modalities.

\paragraph{Table description}
While some HTML tables include captions, the majority do not. To address this, we used an open-source large language model to generate descriptions for all tables, using the raw HTML of the tables as input. We chose a medium-sized model, \texttt{Llama-3-8B-Instruct}, to minimize costs, as the prompt for some tables can be quite large. The details of the prompt can be found in Appendix~\ref{llama3-prompt}.

\paragraph{Image description}

To generate descriptions for images identified as potential charts, we used a vision-language model, \texttt{GPT-4-turbo}. These images were previously identified as potential charts by the ImageNet classifier (see Appendix~\ref{doc_selection}). We prompted the model with the image content, asking it to determine whether the image is a chart, and if so, to specify the type of chart (data chart or map, according to Wikipedia’s definitions). Finally, the model was asked to extract the most relevant information from the chart in fewer than 200 words, which serves as the image description. The full prompt can be found in Appendix~\ref{app-gpt-4}.

\section{Chart Examples}
\label{chart_examples}
\begin{figure*}[htbp]
    \centering

    \begin{subfigure}[b]{0.33\textwidth}
        \includegraphics[width=\linewidth]{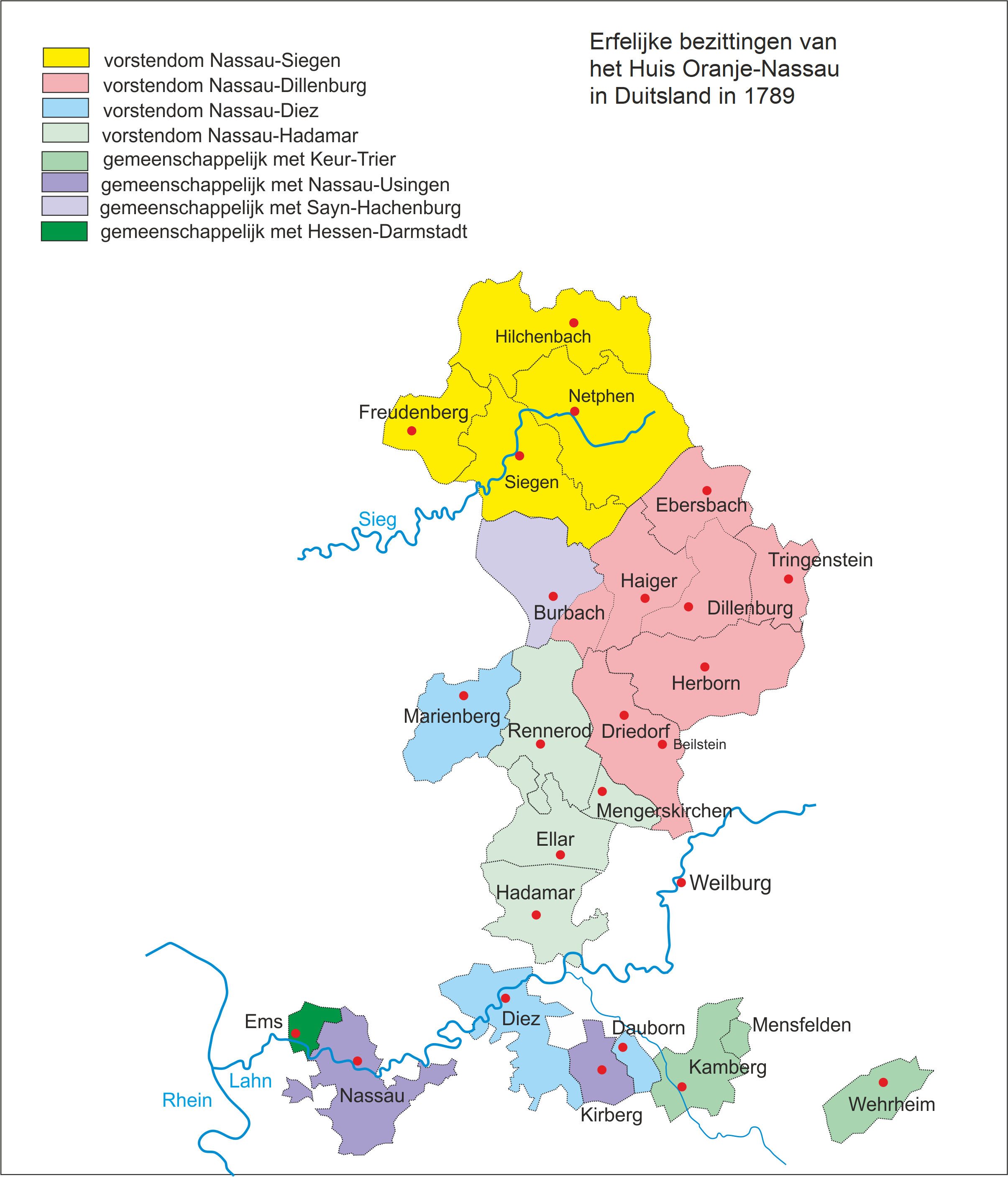}
        \caption{History (Dynasty)}
    \end{subfigure}
    \begin{subfigure}[b]{0.32\textwidth}
        \includegraphics[width=\linewidth]{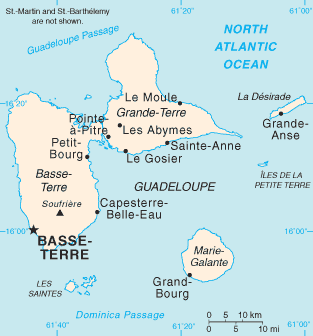}
        \caption{Geography (Region)}
    \end{subfigure}
    \begin{subfigure}[b]{0.3\textwidth}
        \includegraphics[width=\linewidth]{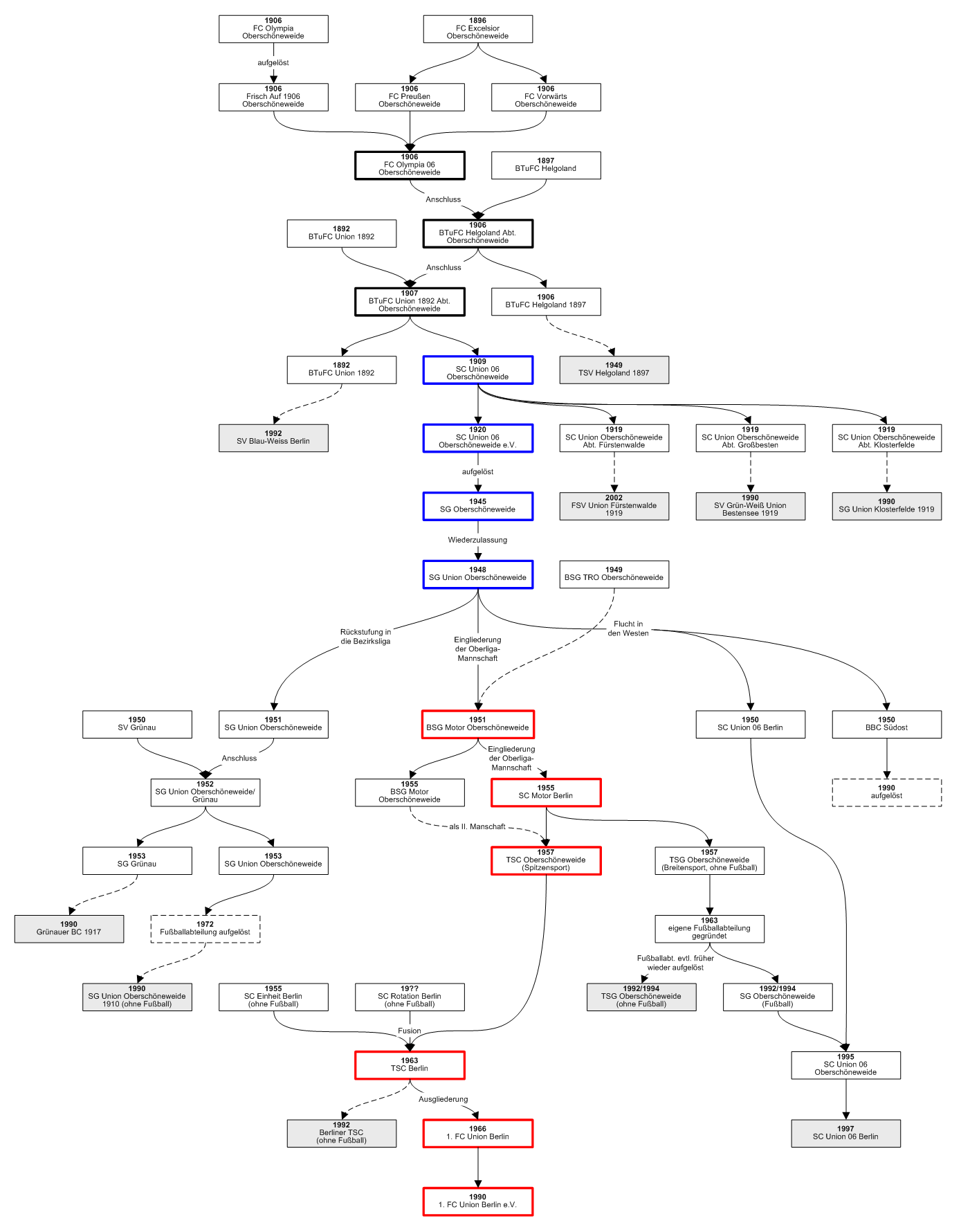}
        \caption{Sport (Teams)}
    \end{subfigure}


    \begin{subfigure}[b]{0.32\textwidth}
        \includegraphics[width=\linewidth]{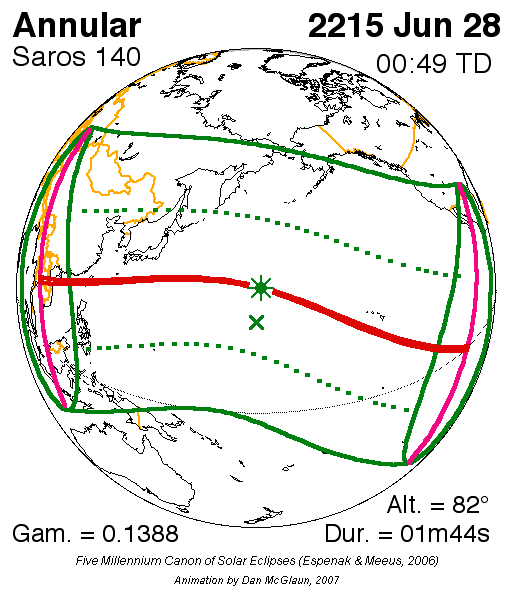}
        \caption{Science (Astronomy)}
    \end{subfigure}
    \begin{subfigure}[b]{0.33\textwidth}
        \includegraphics[width=\linewidth]{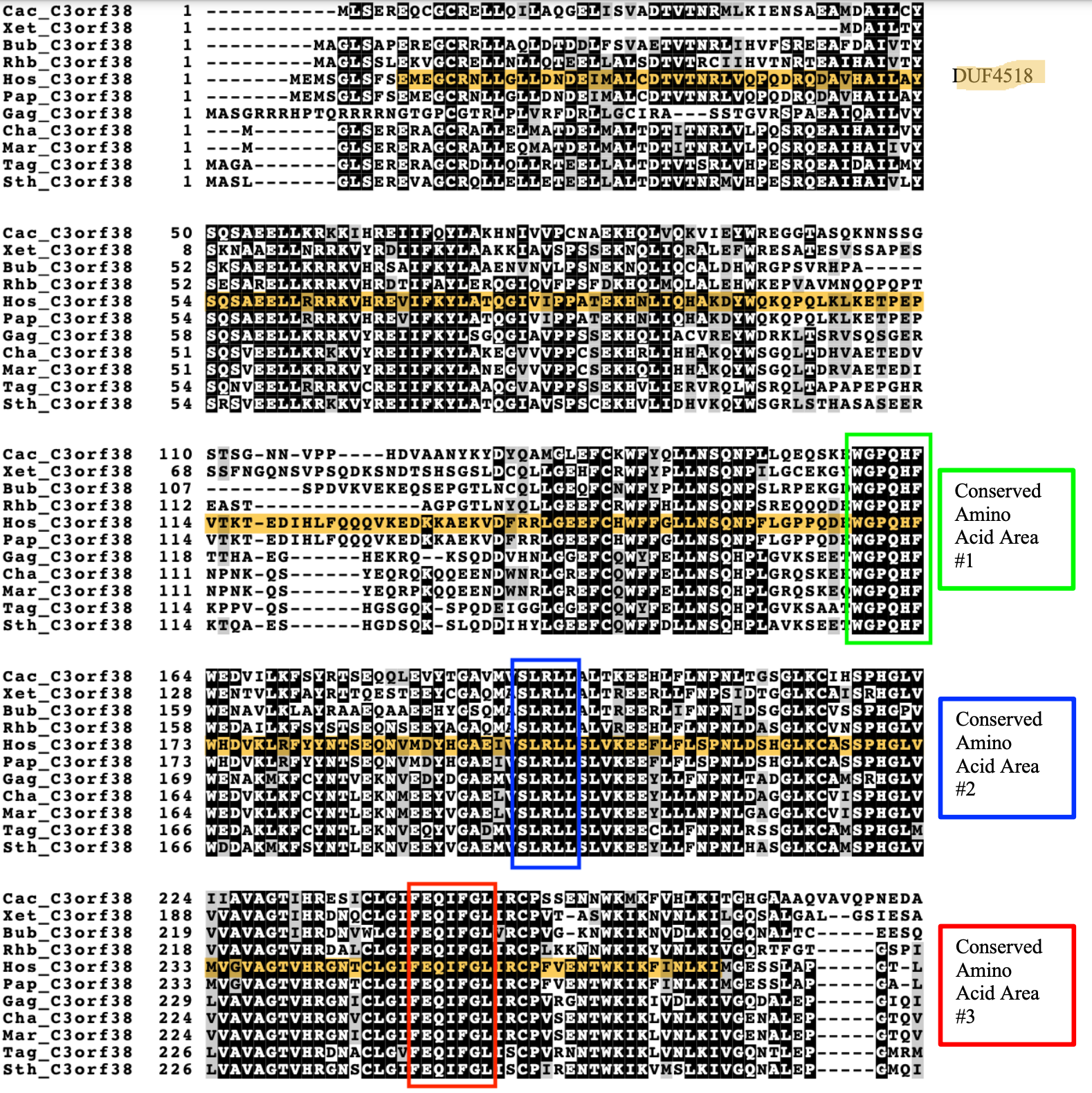}
        \caption{Science (Biology )}
    \end{subfigure}
    \begin{subfigure}[b]{0.33\textwidth}
        \includegraphics[width=\linewidth]{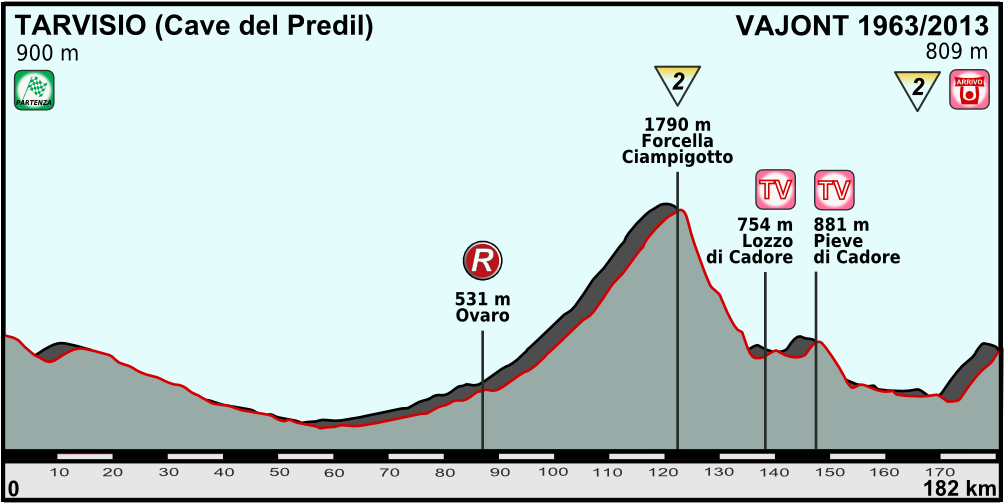}
        \caption{Wikimedia (Article)}
    \end{subfigure}
    
    \begin{subfigure}[b]{1\textwidth}
        \includegraphics[width=\linewidth]{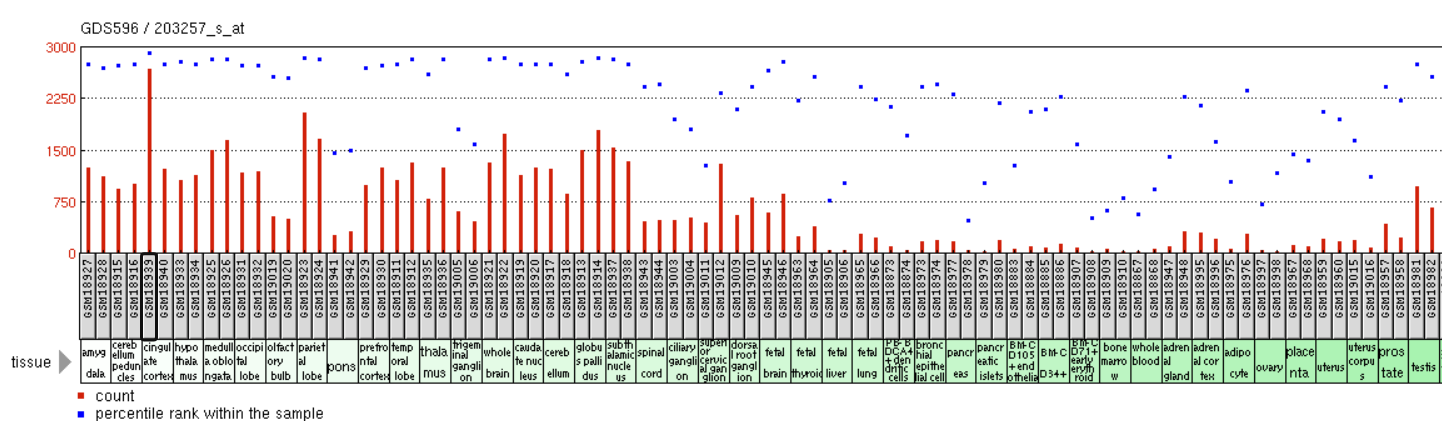}
        \caption{Science (Biology)}
    \end{subfigure}

    \begin{subfigure}[b]{0.33\textwidth}
        \includegraphics[width=\linewidth]{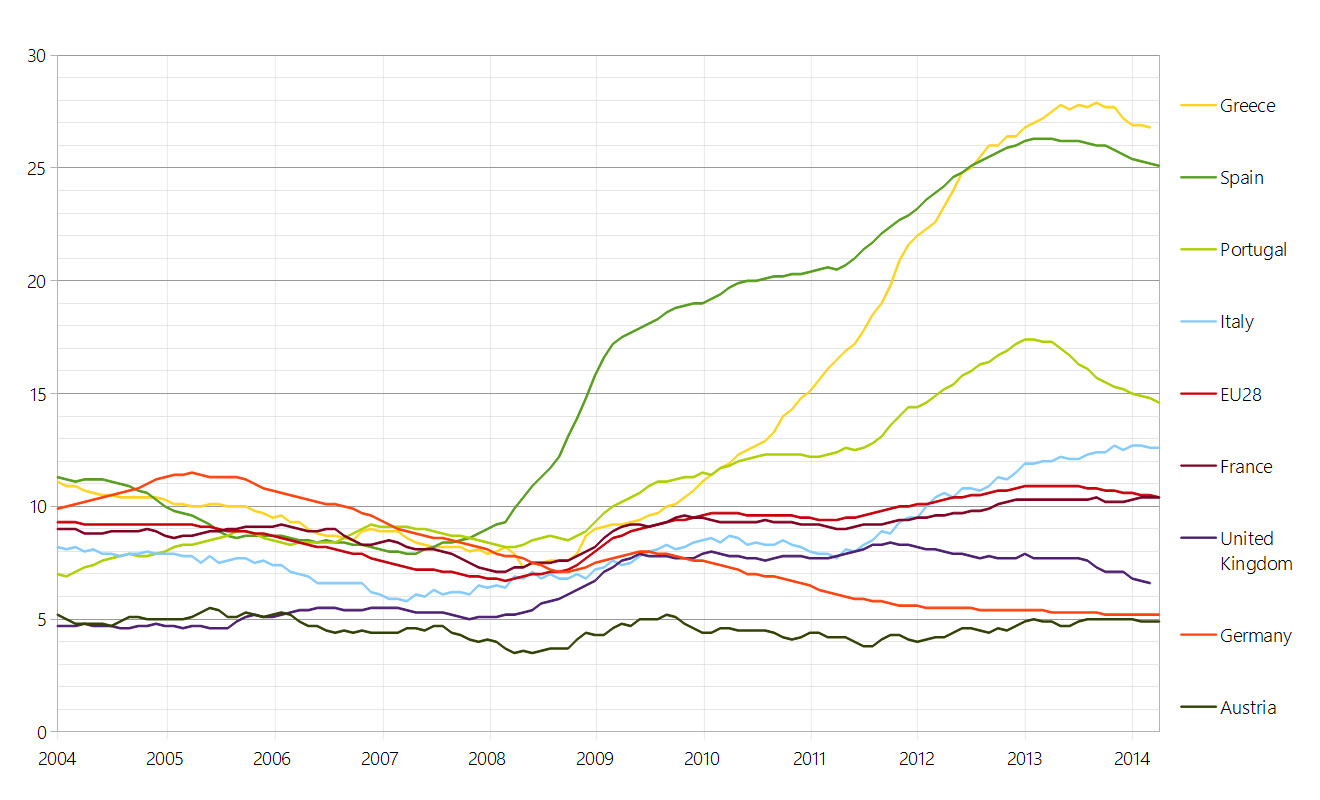}
        \caption{Economy (GDP)}
    \end{subfigure}
    \begin{subfigure}[b]{0.32\textwidth}
        \includegraphics[width=\linewidth]{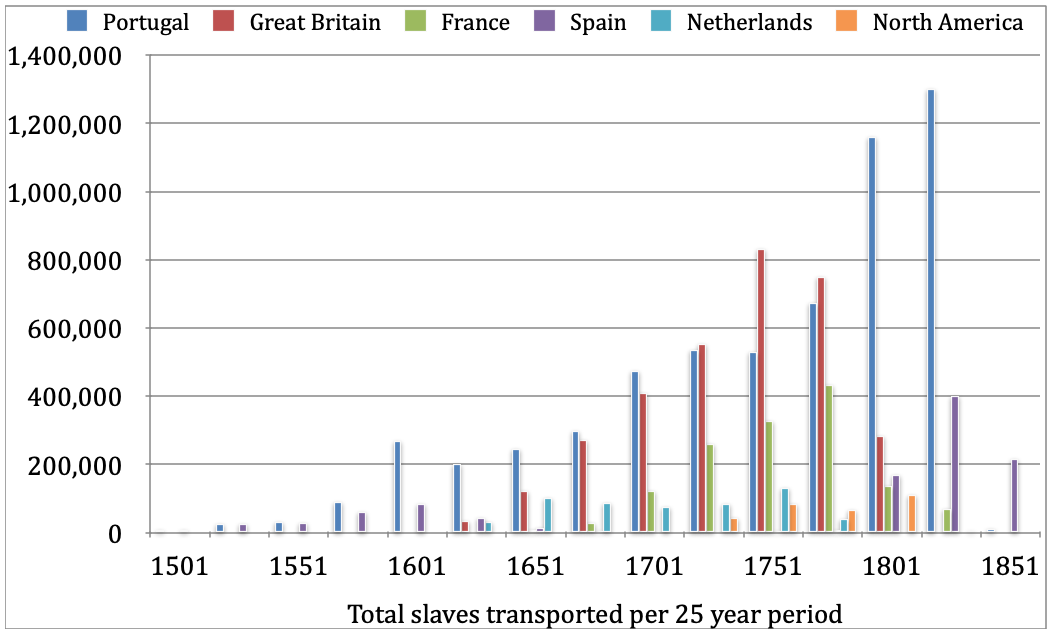}
        \caption{History (Other)}
    \end{subfigure}
    \begin{subfigure}[b]{0.33\textwidth}
        \includegraphics[width=\linewidth]{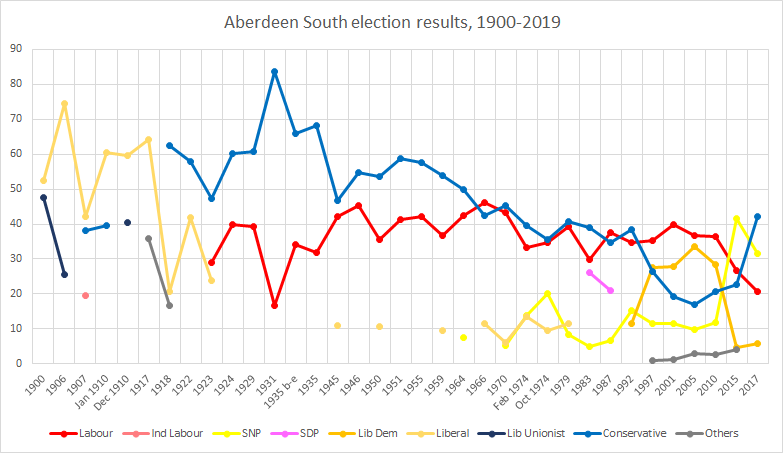}
        \caption{Politics (Composition of parliament, government)}
    \end{subfigure}

    \caption{Chart examples from different topics and subtopics.}
    \label{fig:chart_examples}
\end{figure*}

We define a chart as a graphical representation of data, including: (a) data charts such as diagrams or graphs that organize and display numerical or qualitative information; (b) maps enhanced with additional data; and (c) other domain-specific constructs, such as chord charts or record charts. Figure~\ref{fig:chart_examples} shows some chart examples from different topics and subtopics.
\clearpage
\section{GPT prompts for chart selection}

\subsection{GPT-3.5 Prompt}
\label{app-gpt-3-5}
We used the GPT-3.5-turbo to predict whether an image is likely to represent a chart based on the filename. 

\begin{lstlisting}
You are an image filename reader and you should guess whether the filename describes an image representing a chart or not. You should answer by yes or no in the JSON format.
\end{lstlisting}

\subsection{GPT-4-Turbo Prompt}
\label{app-gpt-4}

We prompted GPT-4 by providing the model the definition from Wikipedia article about “Chart”: \url{https://en.wikipedia.org/wiki/Chart} which identifies three types of charts (data chart, maps and other).

\begin{lstlisting}
Is the image a chart?
A chart (sometimes known as a graph) is a graphical representation for data visualization, in which "the data is represented by symbols, such as bars in a bar chart, lines in a line chart, or slices in a pie chart". A chart can represent tabular numeric data, functions or some kinds of quality structure and provides different info.
If yes, please specify the type of chart between data chart, maps or other.
A data chart is a type of diagram or graph, that organizes and represents a set of numerical or qualitative data.
Maps that are adorned with extra information (map surround) for a specific purpose are often known as charts, such as a nautical chart or aeronautical chart, typically spread over several map sheets.
Then extract the most relevant information from the chart in less than 200 words.
You are communicating with an API, not a user. Begin all AI responses with the character '{' to produce valid JSON. Here is an example:
{ 
   "chart": "yes",
   "type": "data chart",
   "data": "information extracted from the image"
}
\end{lstlisting}

\section{Llama3 prompt for table selection}
\label{llama3-prompt}

We used \texttt{Llama3-8B-Instruct} with \texttt{torchtune}\footnote{Available at \url{https://github.com/pytorch/torchtune}.} to get a textual description of each HTML table. The following prompt has been used:

\begin{lstlisting}
You are a table-to-text assistant. I'll give you a table in HTML format, please write a textual description of the table content."
\end{lstlisting}
\section{GPT-4-turbo prompt for question generation}
\label{app:gpt4-gen}
Prompt used with \texttt{GPT-4-turbo} model for generating the questions with one chart and one HTML table as modalities:
\begin{lstlisting}
You are given one chart and one table. First design a multiple-choice question based on the given chart (Q1). Then design a multiple-choice question based on the given table (Q2). Finally, combine the information from both chart and table to create a multiple-choice question that can be answered ONLY by combining the information from all the given charts and tables (Q3).
Each question should have 4 options, one of which is the correct answer. Please explain how to reach the correct answer from the given context.
You are communicating with an API, not a user. Begin all AI responses with the character '{' to produce valid JSON. Here is an example:
{  
    "Q1":{
    "Question": "<question>",
     "A": "<option1> ",
     "B": "<option2>",
     "C": "<option3>",
     "D": "<option4>", 
     "Answer": "<correct_option>",
     "Explanation": "<explanation>"
     },
     "Q2":{
    "Question": "<question>",
     "A": "<option1> ",
     "B": "<option2>",
     "C": "<option3>",
     "D": "<option4>", 
     "Answer": "<correct_option>",
     "Explanation": "<explanation>"
     },
     "Q3":{
     "Question": "<question>",
     "A": "<option1> ",
     "B": "<option2>",
     "C": "<option3>",
     "D": "<option4>", 
     "Answer": "<correct_option>",
     "Explanation": "<explanation>"
     }
 }
Tables:
{html_table}
{table description if it is available}

<chart image>
\end{lstlisting}

We modified the first paragraph of the prompt and the modalities attached in the end accordingly when two charts or two tables were provided.

    

    
\section{InternVL2 prompt for question-answer pair evaluation}
\label{app:internvl2}

Prompt used with \texttt{InternVL2} model for evaluating the generated question with one chart and one HTML table as modalities:
\begin{lstlisting}
Image:
<image>

HTML table:
{html_table}

Given the information provided in the image and HTML table, can you answer to the following question: {generated_question}
Answer only by yes or no. 
\end{lstlisting}

We modified the start of the prompt accordingly when two charts or two tables were provided.
\section{Answer Generation Prompts}
\label{ap:answer_generation_prompts}

\paragraph{\textit{Oracle} Setting:}

Prompt used in answer-generation step, where we task models to answer a MCQ given provided chart and table (\textit{table-chart}):
\begin{lstlisting}
Given the following chart and table (in HTML format), which of the following answer is correct? A, B, C or D. Please answer in the format A, B, C or D.

Chart:
<image>

HTML table:
{html_table}

\end{lstlisting}

The prompt used in answer-generation step, where we task models to answer a MCQ given two provided tables (\textit{table-table}):
\begin{lstlisting}
Given the following chart and table (in HTML format), which of the following answer is correct? A, B, C or D. Please answer in the format A, B, C or D.

HTML table_1:
{html_table}

HTML table_2:
{html_table}

\end{lstlisting}

Prompt used in answer-generation step, where we task models to answer a MCQ given two provided charts (\textit{chart-chart}):
\begin{lstlisting}
Given the following chart and table (in HTML format), which of the following answer is correct? A, B, C or D. Please answer in the format A, B, C or D.

Chart_1:
<image>

Chart_2:
<image>
\end{lstlisting}

\paragraph{\textit{Wikidoc} Setting:}
The prompt used in answer-generation step, where we task models to answer a MCQ snapshots of the relavant Wikipedia page:

\begin{lstlisting}
Given the following document, which includes tables and charts, which of the following answer is correct? A, B, C or D. 
Extract and analyze the relevant data from the document to select the correct answer. If the document does not contain enough information, infer the most plausible answer or state 'Unable to determine'.
Please answer in the format A, B, C, D, or 'Unable to determine'.

Image:
<image>
.
.
.
Image:
<image>
\end{lstlisting}

\paragraph{\textit{Blind} Setting:}
The prompt used in answer-generation step, where we task models to answer a MCQ given NO ccontextual information:

\begin{lstlisting}
Which of the following answer is correct? A, B, C or D. Please answer in the format A, B, C or D.
            If the chart or table is unavailable, please infer the most plausible answer based on general reasoning or state 'Unable to determine' if no inference is possible. 
            Please answer in the format A, B, C, D, or 'Unable to determine'.
\end{lstlisting}

\begin{figure*}[ht]
    \centering
    \includegraphics[width=\textwidth]{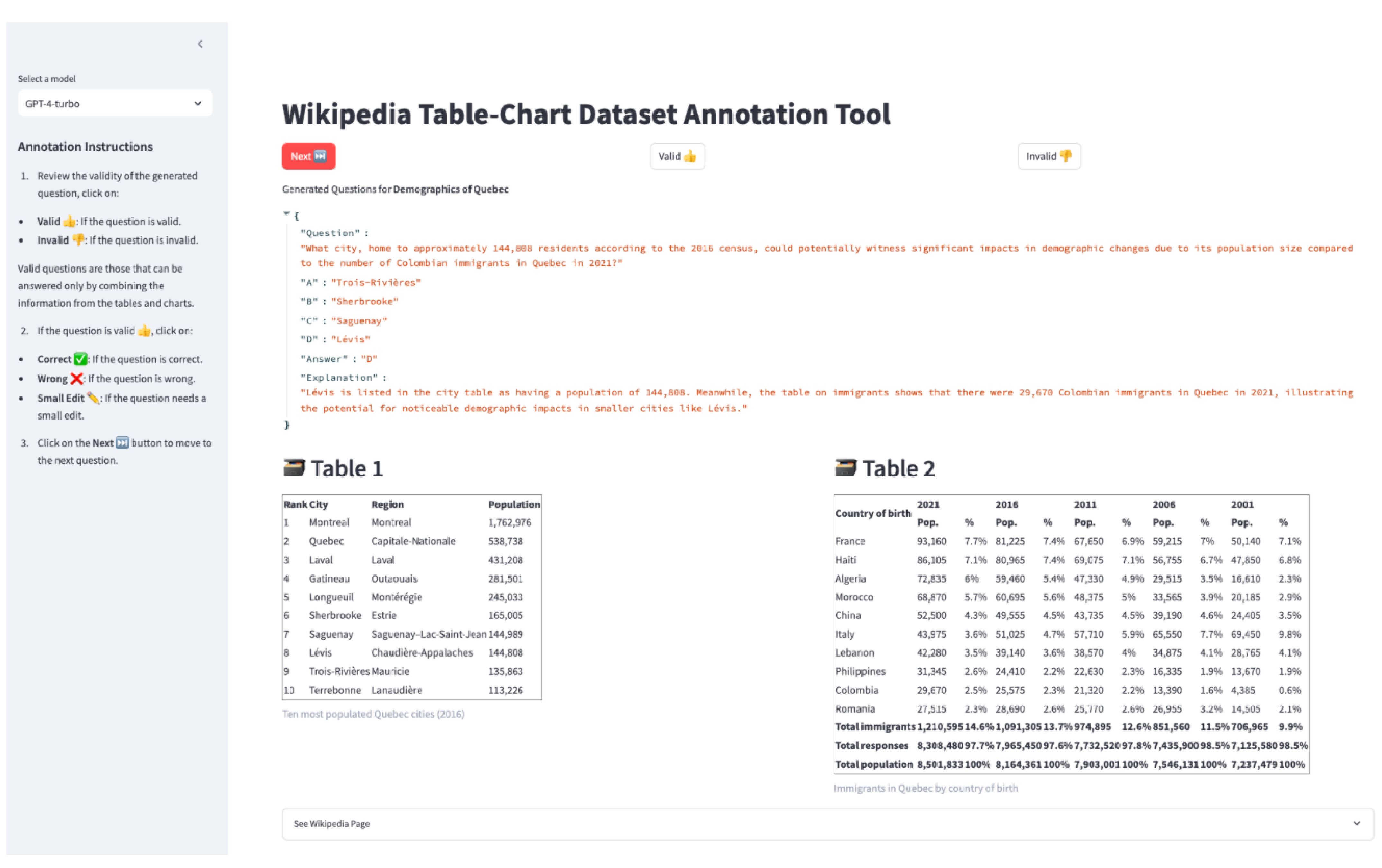}
    \caption{Interface of the annotation tool used for human curation.} 
    \label{fig:annotator}
\end{figure*}
\end{document}